%% file: main.tex
\documentclass{article}
\pdfoutput=1
\PassOptionsToPackage{numbers}{natbib}
\usepackage[final]{nips_2017}

\usepackage[utf8]{inputenc} % allow utf-8 input
\usepackage[T1]{fontenc}    % use 8-bit T1 fonts
\usepackage{hyperref}       % hyperlinks
\usepackage{url}            % simple URL typesetting
\usepackage{booktabs}       % professional-quality tables
\usepackage{amsfonts}       % blackboard math symbols
\usepackage{nicefrac}       % compact symbols for 1/2, etc.
\usepackage{microtype}      % microtypography
\usepackage{graphicx}
\usepackage{subcaption}

\input{math_macros}

\title{Non-Stationary Spectral Kernels}

\author{
  Sami Remes\\
  \And
  Markus Heinonen\\
  \And
  Samuel Kaski
  \AND \vspace{-1em} \\ 
  Helsinki Institute for Information Technology \\
  Department of Computer Science, Aalto University
}

\begin{document}

\maketitle

\begin{abstract}
We propose non-stationary spectral kernels for Gaussian process regression.
We propose to model the spectral density of a non-stationary kernel function as a mixture of input-dependent Gaussian process frequency density surfaces. We solve the generalised Fourier transform with such a model, and present a family of non-stationary and non-monotonic kernels that can learn input-dependent and potentially long-range, non-monotonic covariances between inputs.
We derive efficient inference using model whitening and marginalized posterior, and show with case studies that these kernels are necessary when modelling even rather simple time series, image or geospatial data with non-stationary characteristics.
\end{abstract}

\section{Introduction}

Gaussian processes are a flexible method for non-linear regression \cite{rasmussen2006}. They define a distribution over functions, and their performance depends heavily on the covariance function that constrains the function values. Gaussian processes interpolate function values by considering the value of functions at other similar points, as defined by the kernel function. Standard kernels, such as the Gaussian kernel, lead to smooth neighborhood-dominated interpolation that is oblivious of any periodic or long-range connections within the input space, and can not adapt the similarity metric to different parts of the input space.

Two key properties of covariance functions are \emph{stationarity} and \emph{monotony}. A stationary kernel $K(x,x') = K(x+a,x'+a)$ is a function only of the distance $x-x'$ and not directly the value of $x$. Hence it encodes an identical similarity notion across the input space, while a monotonic kernel decreases over distance. Kernels that are both stationary and monotonic, such as the Gaussian and Mat\'ern kernels, can encode neither input-dependent function dynamics nor long-range correlations within the input space. Non-monotonic and non-stationary functions are commonly encountered in realistic signal processing \cite{rioul1991}, time series analysis \cite{huang1998}, bioinformatics \cite{Grzegorczyk2008,robinson2009}, and in geostatistics applications \cite{higdon1999,huang2008}.

Recently, several authors have explored kernels that are either non-monotonic or non-stationary. A non-monotonic kernel can reveal informative manifolds over the input space by coupling distant points due to periodic or other effects. Non-monotonic kernels have been derived from the Fourier decomposition of kernels \cite{lazaro2010sparse,sinha2016,wilson2013}, which renders them inherently stationary. Non-stationary kernels, on the other hand, are based on generalising monotonic base kernels, such as the Mat\'ern family of kernels \cite{heinonen2016,paciorek2004}, by partitioning the input space \cite{gramacy2008}, or by input transformations \cite{snoek2014}. %Non-stationary kernels can encode different dynamics for different parts of the input space. 

We propose an expressive and efficient kernel family that is -- in contrast to earlier methods -- both non-stationary and non-monotonic, and hence can infer long-range or periodic relations in an input-dependent manner. We derive the kernel from first principles by solving the more expressive \emph{generalised} Fourier decomposition of non-stationary functions, than the more limited standard Fourier decomposition exploited by earlier works. We propose and solve the generalised spectral density as a mixture of Gaussian process density surfaces that model flexible input-dependent frequency patterns. The kernel reduces to a stationary kernel with appropriate parameterisation. We show the expressivity of the kernel with experiments on time series data, image-based pattern recognition and extrapolation, and on climate data modelling.

\section{Non-stationary spectral kernels}\label{sec:model}

This section introduces the main contributions. We employ the generalised spectral decomposition of non-stationary functions and derive a practical and efficient family of kernels based on non-stationary spectral components. Our approach relies on associating input-dependent frequencies for data inputs, and solving a kernel through the generalised spectral transform.

The most general family of kernels is the non-stationary kernels, which include stationary kernels as special cases \cite{genton2001}. 
A non-stationary kernel $k(x,x') \in \R$ for scalar inputs $x,x' \in \R$ can be characterized by its spectral density $S(s,s')$ over frequencies $s,s' \in \R$, and the two are related via a generalised Fourier transform\footnote{We focus on scalar inputs and frequencies for simplicity. An extension based on vector-valued inputs and frequencies \cite{genton2001,kakihara1985} is straightforward.}
\begin{align}
    k(x,x') = \int_{\R} \int_{\R} e^{2\pi i (xs - x' s')} \mu_S(ds, ds') \; , \label{eq:fourier}
\end{align}
where $\mu_S$ is a Lebesgue-Stieltjes measure associated to some positive semi-definite (PSD) spectral density function $S(s,s')$ with bounded variations \cite{genton2001,loeve1978probability,yaglom1987correlation}, which we denote as the \emph{spectral surface} since it considers the amplitude of frequency pairs (See Figure \ref{fig:bsm}a).

The generalised Fourier transform \eqref{eq:fourier} specifies that a spectral surface $S(s,s')$ generates a PSD kernel $K(x,x')$ that is non-stationary unless the spectral measure mass is concentrated only on the diagonal $s=s'$. We design a practical, efficient and flexible parameterisation of spectral surfaces that, in turn, specifies novel non-stationary kernels with input-dependent characteristics and potentially long-range non-monotonic correlation structures.

\subsection{Bivariate Spectral Mixture kernel}
\label{sec:bivariate}

\begin{figure}[t]
    \centering
    \begin{subfigure}[b]{0.49\textwidth}
    \includegraphics[width=\textwidth]{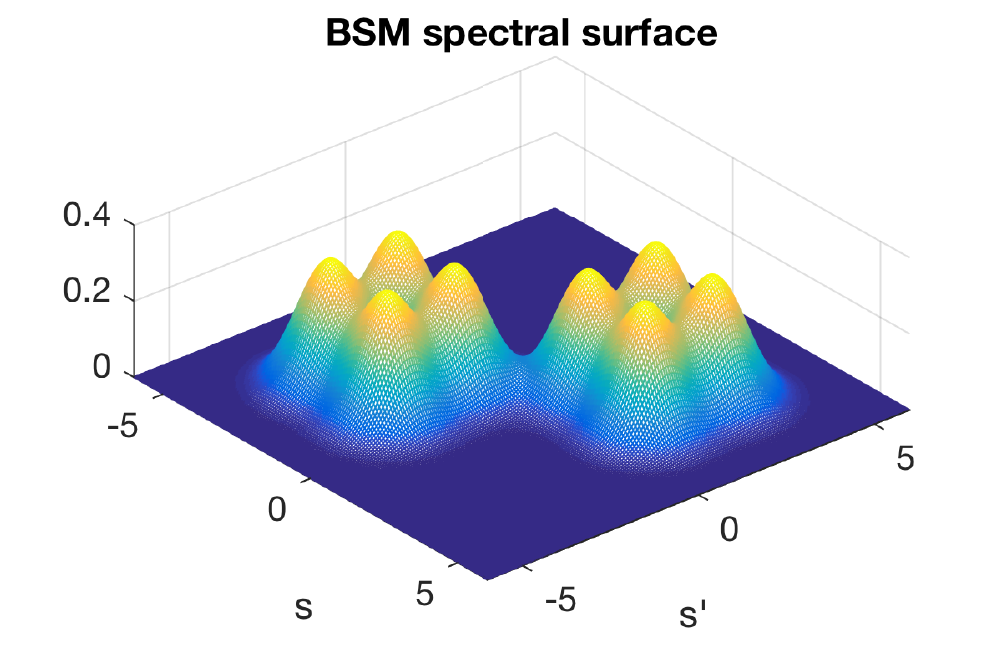}
    \caption{}
    \end{subfigure}    
    \begin{subfigure}[b]{0.49\textwidth}
    \includegraphics[width=\textwidth]{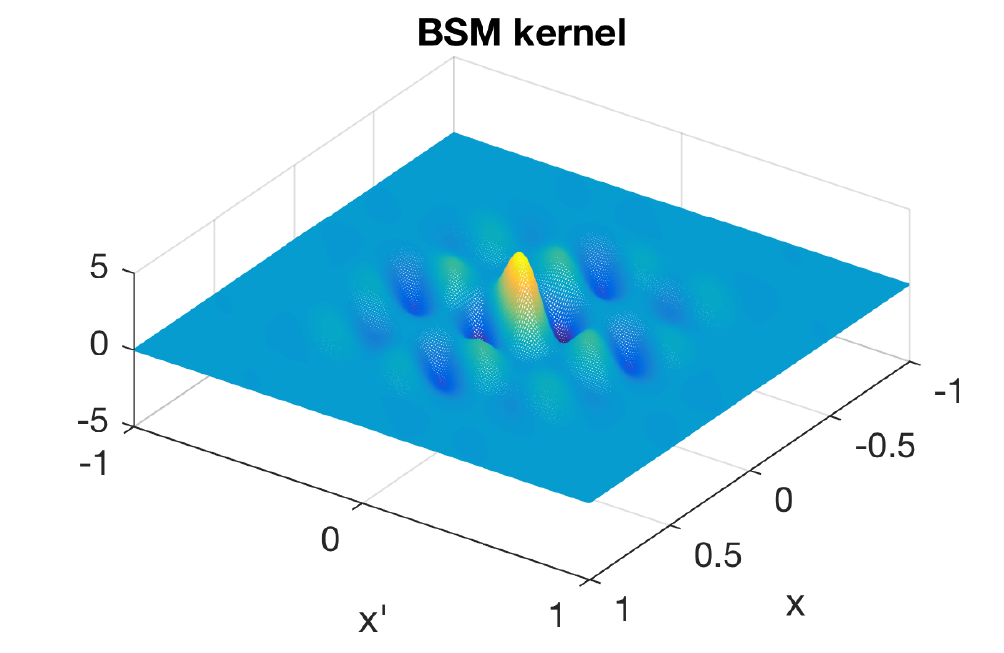}
    \caption{}
    \end{subfigure}

    \caption{\textbf{(a)}: Spectral density surface of a single component bivariate spectral mixture kernel with $8$ permuted peaks. \textbf{(b)}: The corresponding kernel on inputs $x \in [-1,1]$.}
    \label{fig:bsm}
\end{figure}

Next, we introduce spectral kernels that remove the restriction of stationarity of earlier works. We start by modeling the spectral density as a mixture of $Q$ bivariate Gaussian components
\begin{align}
  S_i(s,s') = \sum_{ \bmu_i \in \pm \{ \mu_i, \mu_i'\}^2 } \N \left( \begin{pmatrix} s \\ s' \end{pmatrix} | \bmu_i, \Sigma_i\right), \qquad \Sigma_i = \begin{bmatrix} \sigma_i^2 & \rho_i \sigma_i \sigma_i' \\ \rho_i \sigma_i \sigma_i' & {\sigma_i'}^2 \end{bmatrix} \label{eq:spect} \; ,
\end{align}
with parameterization using the correlation $\rho_i$, means $\mu_i,\mu_i'$ and variances $\sigma_i^2,{\sigma_i'}^2$. To produce a PSD spectral density $S_i$ as required by equation \eqref{eq:fourier} we need to include symmetries $S_i(s,s') = S_i(s',s)$ and sufficient diagonal components $S_i(s,s)$, $S_i(s',s')$. To additionally result in a real-valued kernel, symmetry is required with respect to the negative frequencies as well, i.e., $S_i(s,s') = S_i(-s,-s')$. The sum $\sum_{ \bmu_i \in \pm \{ \mu_i, \mu_i'\}^2 }$ satisfies all three requirements by iterating over the four permutations of $\{\mu_i,\mu_i'\}^2$ and the opposite signs $(-\mu_i, -\mu_i')$, resulting in eight components (see Figure \ref{fig:bsm}a).

The generalised Fourier transform \eqref{eq:fourier} can be solved in closed form for a weighted spectral surface mixture $S(s,s') = \sum_{i=1}^Q w_i^2 S_i(s,s')$ using Gaussian integral identities (see the appendix):
\begin{align}
k(x,x') %&= \iint S(s,s') e^{2\pi i (xs - x' s')} ds ds' \notag \\
 &= \sum_{i=1}^Q w_i^2 \exp(- 2 \pi^2 \tilde{\x}^T \Sigma_i \tilde{\x}) \Psi_{\mu_i,\mu'_i}(x)^T\Psi_{\mu_i,\mu'_i}(x') \label{eq:wm}
 \end{align}
where
\begin{align*}
\Psi_{\mu_i,\mu'_i}(x) &= \begin{pmatrix} \cos 2\pi\mu x + \cos 2\pi\mu' x \\ \sin 2\pi\mu x + \sin 2\pi\mu' x \end{pmatrix},
\end{align*}
and where we define $\tilde{\x} = ( x, -x')^T$ and introduce mixture weights $w_i$ for each component. We denote the proposed kernel as the \emph{bivariate spectral mixture} (BSM) kernel (see Figure \ref{fig:bsm}b). The positive definiteness of the kernel is guaranteed by the spectral transform, and is also easily verified since the sinusoidal components form an inner product and the exponential component resembles an unscaled Gaussian density.

We immediately notice that the BSM kernel vanishes rapidly outside the origin $(x,x') = (0,0)$. We would require a huge number of components centered at different points $x_i$ to cover a reasonably-sized input space.

\subsection{Generalised Spectral Mixture (GSM) kernel}

\begin{figure}[t]
    \centering
    \begin{subfigure}[b]{0.245\textwidth}
    \includegraphics[width=\textwidth]{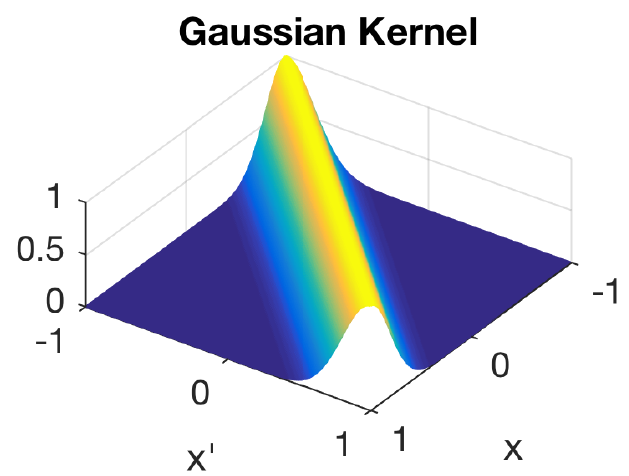}
    \caption{}
    \end{subfigure}
    \begin{subfigure}[b]{0.245\textwidth}
    \includegraphics[width=\textwidth]{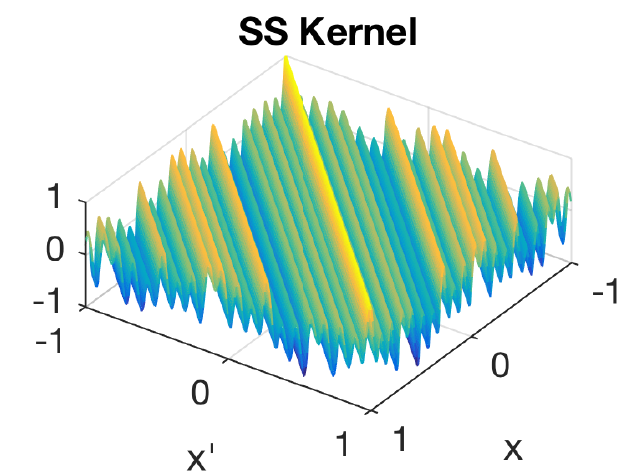}
    \caption{}
    \end{subfigure}
    \begin{subfigure}[b]{0.245\textwidth}
    \includegraphics[width=\textwidth]{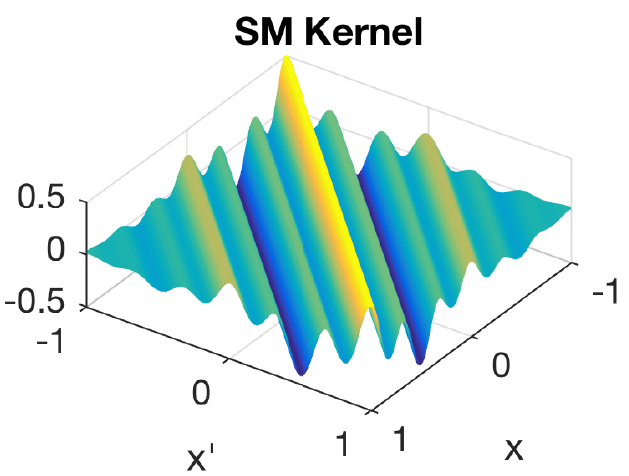}
    \caption{}
    \end{subfigure}
    \begin{subfigure}[b]{0.245\textwidth}
    \includegraphics[width=\textwidth]{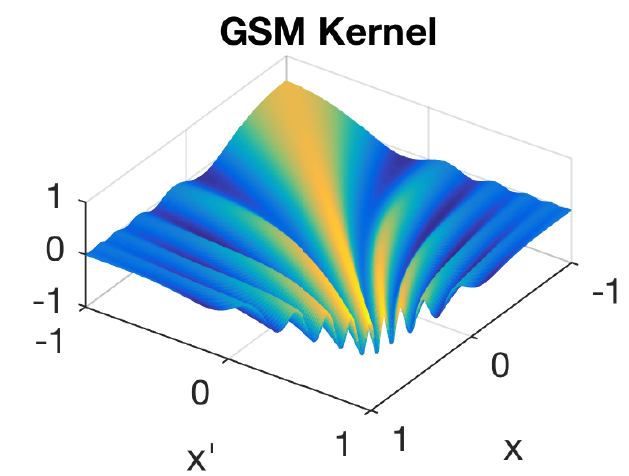}
    \caption{}
    \end{subfigure}

    \begin{subfigure}[b]{0.245\textwidth}
    \includegraphics[width=\textwidth]{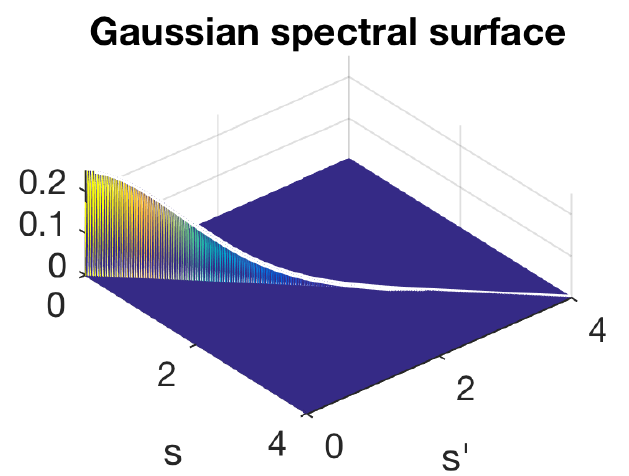}
    \caption{}
    \end{subfigure}
    \begin{subfigure}[b]{0.245\textwidth}
    \includegraphics[width=\textwidth]{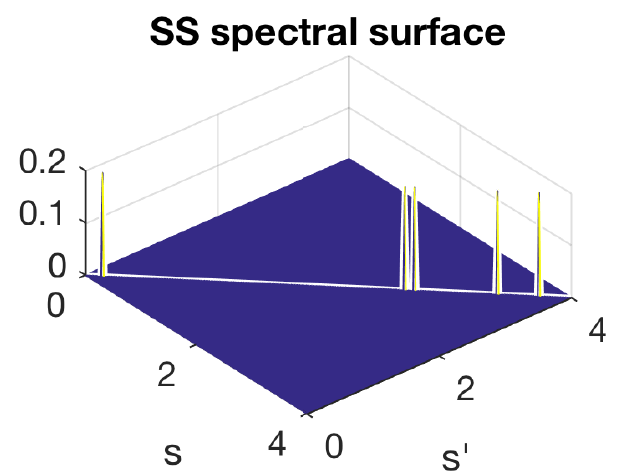}
    \caption{}
    \end{subfigure}
    \begin{subfigure}[b]{0.245\textwidth}
    \includegraphics[width=\textwidth]{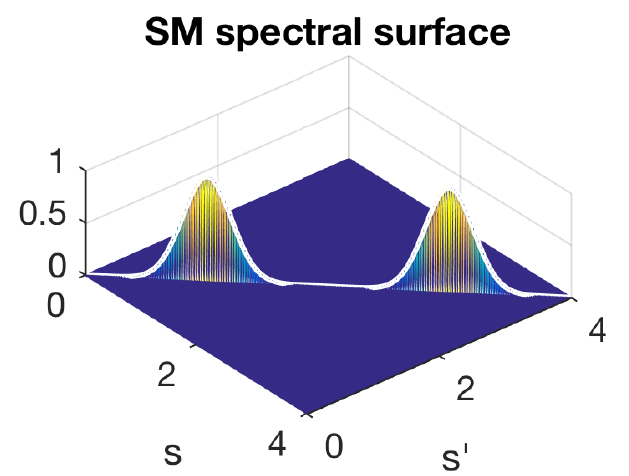}
    \caption{}
    \end{subfigure}
    \begin{subfigure}[b]{0.245\textwidth}
    \includegraphics[width=\textwidth]{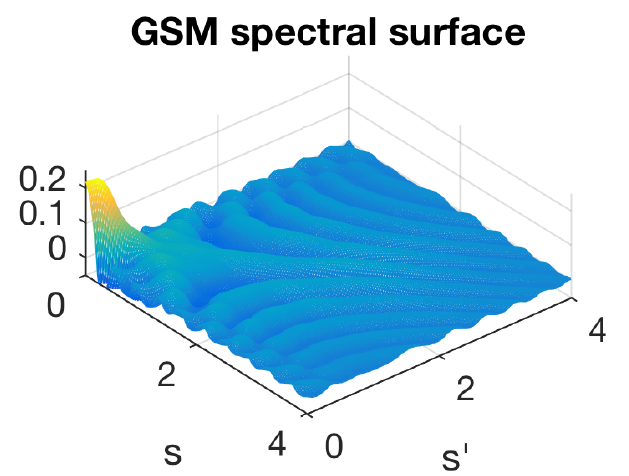}
    \caption{}
    \end{subfigure}

    \begin{subfigure}[b]{0.245\textwidth}
    \includegraphics[width=\textwidth]{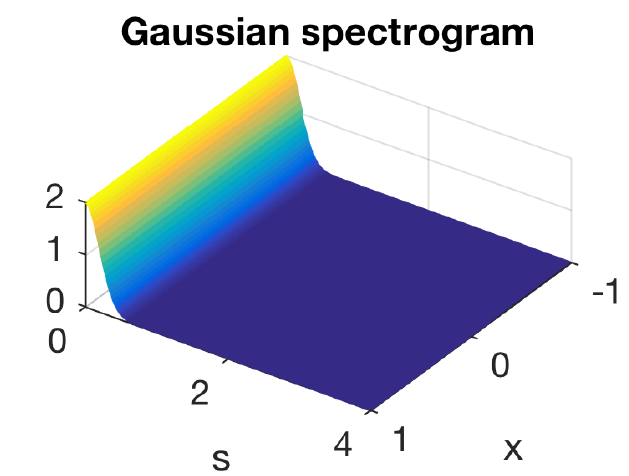}
    \caption{}
    \end{subfigure}
    \begin{subfigure}[b]{0.245\textwidth}
    \includegraphics[width=\textwidth]{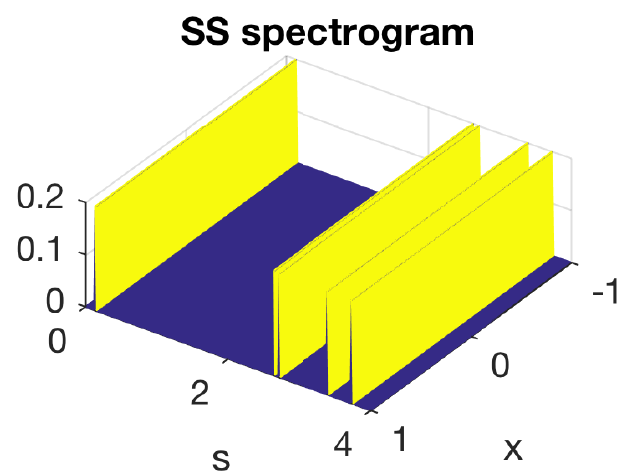}
    \caption{}
    \end{subfigure}
    \begin{subfigure}[b]{0.245\textwidth}
    \includegraphics[width=\textwidth]{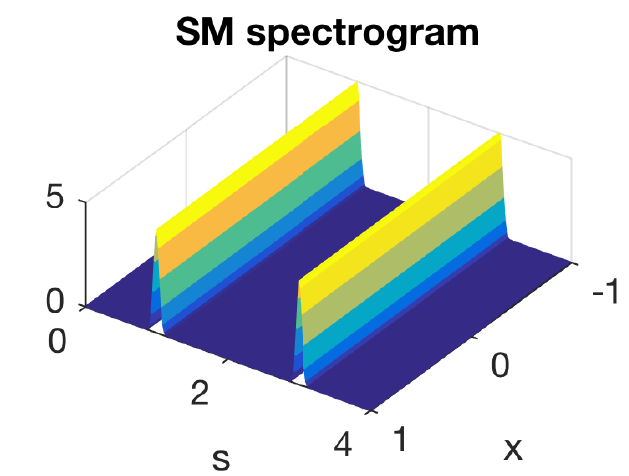}
    \caption{}
    \end{subfigure}
    \begin{subfigure}[b]{0.245\textwidth}
    \includegraphics[width=\textwidth]{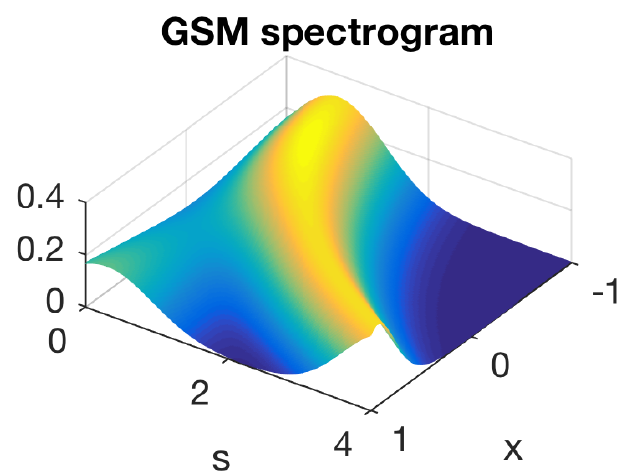}
    \caption{}
    \end{subfigure}

    \caption{\textbf{(a)-(d)}: Examples of kernel matrices on inputs $x \in [-1,1]$ for a Gaussian kernel (a), sparse spectrum kernel \cite{lazaro2010sparse} (b), spectral mixture kernel \cite{wilson2013} (c), and for the GSM kernel (d). \textbf{(e)-(h)}: The corresponding generalised spectral density surfaces of the four kernels. \textbf{(i)-(l)}: The corresponding spectrograms, that is, input-dependent frequency amplitudes. The GSM kernel is highlighted with a spectrogram mixture of $Q=2$ Gaussian process surface functions.}
    \label{fig:surfaces}
\end{figure}

To overcome the deficiencies of the kernel derived in Section~\ref{sec:bivariate}, we extend it further by parameterizing the frequencies, length-scales and mixture weights as a Gaussian processes\footnote{See the appendix for a tutorial on Gaussian processes.}, that form a smooth spectrogram (See Figure \ref{fig:surfaces}l):
\begin{align}
    \log w_i(x) \sim \GP(0,k_w(x,x')), \\
    \log \ell_i(x) \sim \GP(0,k_\ell(x,x')), \\
    \logit \mu_i(x) \sim \GP(0,k_\mu(x,x')).
\end{align}
Here the log transform is used to ensure the weights $w(x)$ and lengthscales $\ell(x)$ are non-negative, and the logit transform $\logit \mu(x) = \log\frac{\mu}{F_N-\mu}$ limits the learned frequencies between zero and the Nyquist frequency $F_N$, which is defined as half of the sampling rate of the signal. %(or for non-equispaced signals as the inverse of the smallest time interval between the samples).

A GP prior $f(x) \sim \GP(0, k(x,x'))$ defines a distribution over zero-mean functions, and denotes the covariance between function values $\cov[ f(x), f(x')] = k(x,x')$ as the prior kernel For any collection of inputs, $x_1, \ldots, x_N$, the function values follow a multivariate normal distribution $(f(x_1), \ldots, f(x_N))^T \sim \N(\0, K)$, where $K_{ij} = k(x_i, x_j)$. The key property of Gaussian processes is that they can encode smooth functions by correlating function values of input points that are similar according to the  kernel $k(x,x')$. We use standard Gaussian kernels $k_w$, $k_\ell$ and $k_\mu$.

We accommodate the input-dependent lengthscale by replacing the exponential part of \eqref{eq:wm} by the Gibbs kernel
\begin{align*}
    k_{\text{Gibbs},i}(x,x') = \sqrt{\frac{2\ell_i(x) \ell_i(x')}{\ell_i(x)^2+\ell_i(x')^2}}\exp\left(-\frac{(x-x')^2}{\ell_i(x)^2+\ell_i(x')^2}\right) \; ,
\end{align*}
which is a non-stationary generalisation of the Gaussian kernel \cite{gibbs1997,heinonen2016,paciorek2004}. 
We propose a non-stationary \emph{generalised spectral mixture} (GSM) kernel with a simple closed form (see the appendix):
\begin{align}
    k_{\text{GSM}}(x,x') = \sum_{i=1}^Q w_i(x) w_i(x') k_{gibbs,i}(x,x') \cos(2 \pi (\mu_i(x) x - \mu_i(x') x')) \; . \label{eq:gsm}
\end{align}
The kernel is a product of three PSD terms. The GSM kernel encodes the similarity between two data points based on their combined signal variance $w(x)w(x')$, and the frequency surface based on the frequencies $\mu(x),\mu(x')$ and frequency lengthscales $\ell(x),\ell(x')$ associated with both inputs. The GSM kernel encodes the spectrogram surface mixture into a relatively simple kernel. The kernel reduces to the stationary Spectral Mixture (SM) kernel \cite{wilson2013} with constant functions $w_i(x) = w_i$, $\mu_i(x) = \mu_i$ and $\ell_i(x) = 1/(2\pi \sigma_i)$ (see the appendix).

We have presented the proposed kernel \eqref{eq:gsm} for univariate inputs for simplicity. The kernel can be extended to multivariate inputs in a straightforward manner using the generalised Fourier transform with vector-valued inputs \cite{genton2001,kakihara1985}. However, since in many applications multivariate inputs have a grid-like structure, for instance in geostatistics, image analysis and temporal models. We exploit this assumption and propose a multivariate extension that assumes the inputs to decompose across input dimensions \cite{flaxman2015,wilson2013}:
\begin{align}
k_{\text{GSM}}(\x,\x' | \boldsymbol\theta) = \prod_{p=1}^P k_{\text{GSM}}(x_p, x_p' | \bt_p) \; .
\end{align}
Here $\x,\x' \in \R^P$, $\bt = (\bt_1, \ldots, \bt_P)$ collects the dimension-wise kernel parameters $\bt_p = (\w_{ip}, \bl_{ip}, \bmu_{ip})_{i=1}^Q$ of the $N$-dimensional realisations $\w_{ip},\bl_{ip},\bmu_{ip} \in \R^N$ per dimension $p$. Then, the kernel matrix can be expressed using Kronecker products as $\K_\bt = K_{\bt_1} \otimes \cdots \otimes K_{\bt_P}$, while missing values and data not on a regular grid can be handled with standard techniques \cite{flaxman2015,saatcci2011scalable,wilson2015kissgp}.

\section{Inference}

We use the Gaussian process regression framework and assume a Gaussian likelihood over $N^P$ data points $(\x_j,y_j)_{j=1}^{N^P}$ with all outputs collected into a vector $\y \in \R^{N^P}$,
\begin{align}
y_j &= f(\x_j) + \varepsilon_j, \qquad \varepsilon_j \sim \N(0, \sigma_n^2) \notag \\
 f(\x) &\sim \GP(0, k_{\text{GSM}}(\x,\x' | \bt)),
\end{align}
with a standard predictive GP posterior $f(\x_\star | \y)$ for a new input point $\x_\star$ \cite{rasmussen2006}. The posterior can be efficiently computed using Kronecker identities \cite{saatcci2011scalable} (see the appendix).

We aim to infer the noise variance $\sigma_n^2$ and the kernel parameters $\boldsymbol\theta = (\w_{ip}, \bl_{ip}, \bmu_{ip})_{i=1,p=1}^{Q,P}$ that reveal the input-dependent frequency-based correlation structures in the data, while regularising the learned kernel to penalise overfitting. We perform MAP inference over the log marginalized posterior $\log p(\bt | \y) \propto \log p(\y | \bt) p(\bt) = \mathcal{L}(\bt)$, where the functions $f(x)$ have been marginalised out,
\begin{align}
\mathcal{L}(\bt) &= \log \left( \N(\y | \0, \K_{\bt} + \sigma_n^2\I) \prod_{i,p=1}^{Q,P} \N(\w_{ip} | \0, K_{w_p}) \N(\bmu_{ip} | \0, K_{\mu_p}) \N(\bl_{ip} | \0, K_{\ell_p}) \right), \label{eq:gpmll}
\end{align}
where $K_{w_p},K_{\mu_p},K_{\ell_p}$ are $N \times N$ prior matrices per dimensions $p$. The marginalized posterior automatically balances between parameters $\bt$ that fit the data and a model that is not overly complex \cite{rasmussen2006}. 
We can efficiently evaluate both the marginalized posterior and its gradients in $\mathcal{O}( P N^{\frac{P+1}{P}} )$ instead of the usual $\mathcal{O}( {N^P}^3 )$ complexity \cite{saatcci2011scalable} (see the appendix).

Gradient-based optimisation of \eqref{eq:gpmll} is likely to converge very slowly due to parameters $\w_{ip}, \bmu_{ip}, \bl_{ip}$ being highly self-correlated. We remove the correlations by whitening the variables as $\tilde\bt = \LL^{-1}\bt$ where $\LL$ is the Cholesky decomposition of the prior covariances. We maximize $\L(\bt)$ using gradient ascent with respect to the whitened variables $\tbt$ by evaluating $\L( \LL \tbt)$ and the gradient as \cite{heinonen2016,kuss2005}
\begin{align}
\pdd{\L(\bt)}{\tbt} = \pdd{\L(\bt)}{\bt}\pdd{\bt}{\tbt} = \LL^T \pdd{\L(\bt)}{\bt}. \label{eq:grad}
\end{align}

\section{Related Work}\label{sec:related}

Bochner's theorem for stationary signals, whose covariance can be written as $k(\tau) = k(x-x') = k(x,x')$, implies a Fourier dual \cite{wilson2013}
\begin{align*}
k(\tau) &= \int S(s) e^{2\pi i s \tau} ds \\
S(s) &= \int k(\tau) e^{-2\pi i s \tau} d\tau.
\end{align*}
The dual is a special case of the more general Fourier transform \eqref{eq:fourier}, and has been exploited to design rich, yet stationary kernel representations \cite{sinha2016,yang2015} and used for large-scale inference \cite{rahimi2008}. 
Lazaro-Gredilla et al.~proposed to directly learn the spectral density as a mixture of Dirac delta functions leading to a sparse spectrum (SS) kernel $k_{\text{SS}}(\tau) = \frac{1}{Q} \sum_{i=1}^Q \cos(2 \pi s_i^T \tau)$ \cite{lazaro2010sparse}. Wilson et al.~derived a stationary spectral mixture (SM) kernel by modelling the univariate spectral density using a mixture of normals $S_{\text{SM}}(s) = \sum_i w_i [\N(s|\mu_i,\sigma_i^2) + \N(s| -\mu_i,\sigma_i^2)] / 2$ \cite{wilson2013}, corresponding to the kernel function $k_\text{SM}(\tau) = \sum_i w_i\exp(-2\pi^2\sigma_i^2\tau)\cos(2\pi\mu_i\tau)$, which we generalized to the non-stationary case. Kernels derived from the spectral representation are particularly well suited to encoding long-range, non-monotonic or periodic kernels; however, they have so far been unable to handle non-stationarity. 

Non-stationary kernels, on the other hand, have been constructed by non-stationary extensions of Mat\'ern and Gaussian kernels with input-dependent lengthscales \cite{gibbs1997,heinonen2016,paciorek2004,paciorek2006}, input space warpings \cite{sampson1992,snoek2014}, and with local stationarity with products of stationary and non-stationary kernels \cite{genton2001,silverman1957}. The simplest non-stationary kernel is arguably the dot product kernel \cite{rasmussen2006}, which has been used as a way to assign input-dependent signal variances \cite{tolvanen2014}. Non-stationary kernels are a good match for functions with transitions in their dynamics, yet are unsuitable for modelling non-monotonic properties.

Our work can also be seen as a generalisation of wavelets, or time-dependent frequency components, into general and smooth input-dependent components. In signal processing, Hilbert-Huang transforms and Hilbert spectral analysis explore input-dependent frequencies, but with deterministic transform functions on the inputs \cite{huang2008,huang1998}.

\section{Experiments}\label{sec:experiments}

We apply our proposed kernel first on simple simulated time series, then on texture images and lastly on a land surface temperature dataset. With the image data, we compare our method to two stationary mixture kernels, specifically the spectral mixture (SM) \cite{wilson2013} and sparse spectrum (SS) kernels \cite{lazaro2010sparse}, and the standard squared exponential (SE) kernel. We employ the GPML Matlab toolbox, which directly implements the SM and SE kernels, and the SS kernel as a meta kernel combining simple cosine kernels. The GPML toolbox also implements Kronecker inference automatically for these kernels. 
We implemented the proposed GSM kernel and inference in Matlab.

For optimizing the log posterior \eqref{eq:gpmll} we employ the L-BFGS algorithm. For both our method and the comparisons, we restart the optimization from 10 different initialisations, each of which is chosen as the best among 100 randomly sampled hyperparameter values as evaluating the log posterior is cheap compared to evaluating gradients or running the full optimisation.

\subsection{Simulated time series with a decreasing frequency component}

\begin{figure}
    \centering
    \begin{subfigure}[b]{0.48\textwidth}
    \centering
    \includegraphics[height=1.6in]{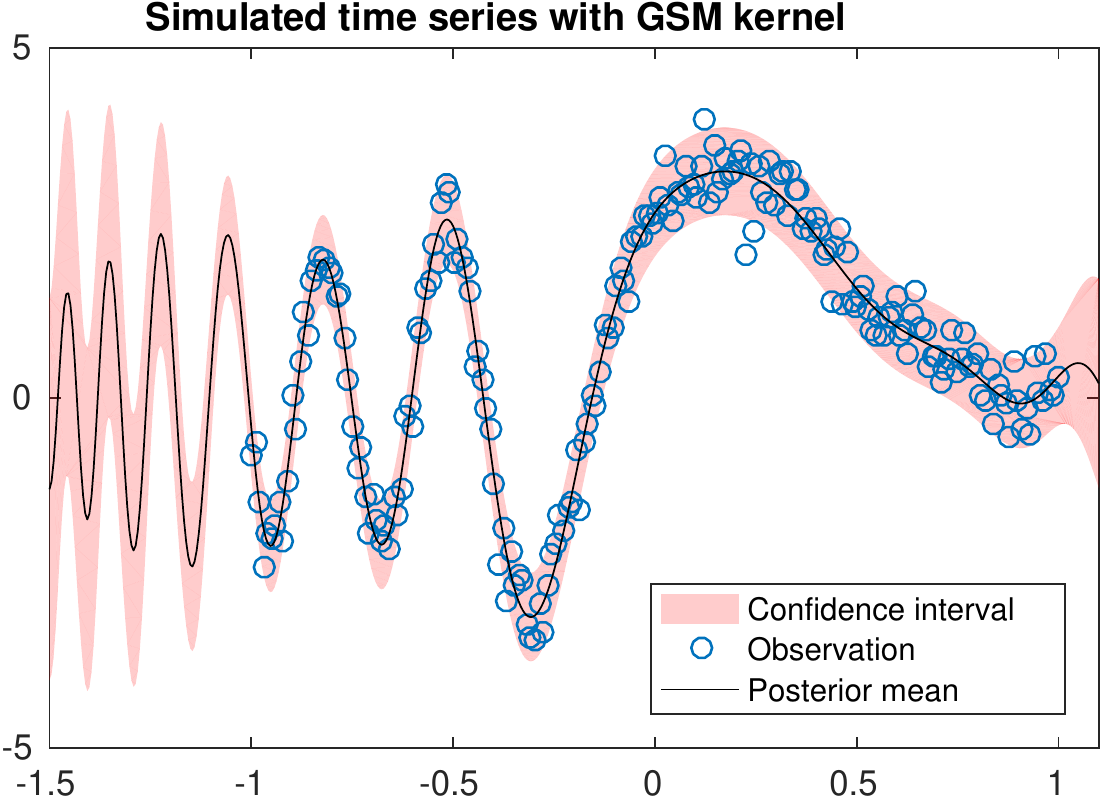}
    \caption{}
    \end{subfigure}
    \begin{subfigure}[b]{0.48\textwidth}
    \centering
    \includegraphics[height=1.7in]{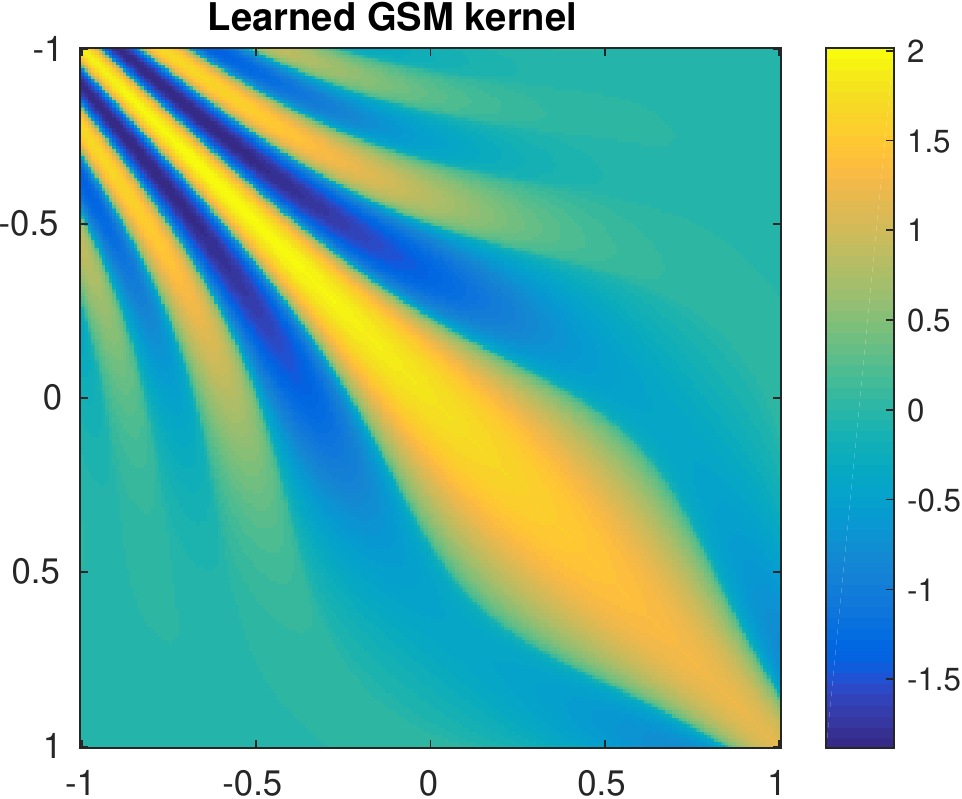}
    \caption{}
    \end{subfigure}
    \caption{\textbf{(a)} A simulated time series with a single decreasing frequency component and a GP fitted using a GSM kernel. \textbf{(b)} The learned kernel shows that close to $x=-1$ the signal is highly correlated and anti-correlated with close time points, while these longer-range dependencies vanish when moving towards $x=1$.}
    \label{fig:incfreq}
\end{figure}

First we experiment whether the GSM kernel can find a simulated time-varying frequency pattern. We simulated a dataset where the frequency of the signal changes deterministically as $\mu(x) = 1+(1-x)^2$ on the interval $x\in[-1,1]$. We built a single-component GSM kernel $K$ using the specified functions $\mu(x)$, $\ell(x) = \ell = \exp(-1)$ and $w(x) = w = 1$. We sampled a noisy function $\y \sim \N(\vec0,K+\sigma_n^2 I)$ with a noise variance $\sigma_n^2 = 0.1$. The example in Figure~\ref{fig:incfreq} shows the learned GSM kernel, as well as the data and the function posterior $f(x)$. For this 1D case, we also employed the empirical spectrogram for initializing the hyperparameter values. The kernel correctly captures the increasing frequency towards negative values (towards left in Figure \ref{fig:incfreq}a). 

\subsection{Image data}

We applied our kernel to two texture images. The first image of a sheet of metal represents a mostly stationary periodic pattern. The second, a wood texture, represents an example of a very non-stationary pattern, especially on the horizontal axis. We use majority of the image as training data (the non-masked regions of Figure \ref{fig:incfreq}a and \ref{fig:incfreq}f) , and use the compared kernels to predict a missing cross-section in the middle, and also to extrapolate outside the borders of the original image.

\begin{figure}[t]
    \centering
    \begin{subfigure}[b]{0.195\textwidth}
    \includegraphics[width=\textwidth]{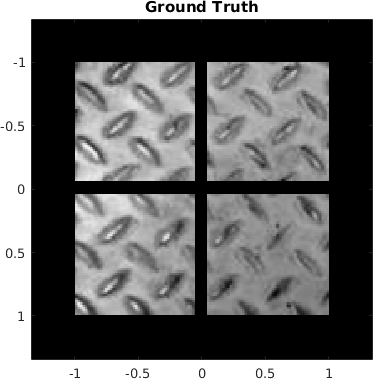}
    \caption{}
    \end{subfigure}
    \begin{subfigure}[b]{0.195\textwidth}
    \includegraphics[width=\textwidth]{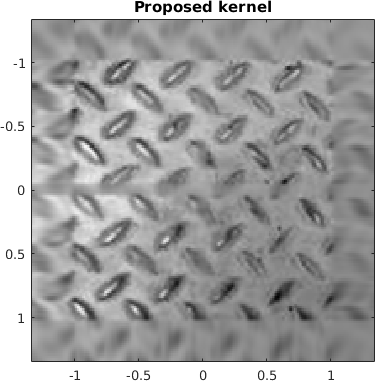}
    \caption{}
    \end{subfigure}
    \begin{subfigure}[b]{0.195\textwidth}
    \includegraphics[width=\textwidth]{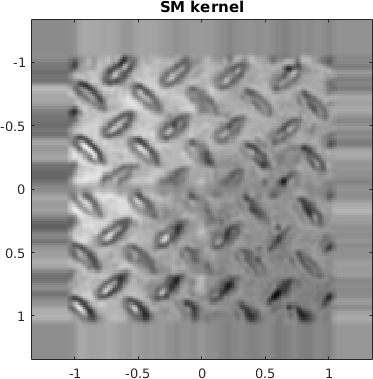}
    \caption{}
    \end{subfigure}
    \begin{subfigure}[b]{0.195\textwidth}
    \includegraphics[width=\textwidth]{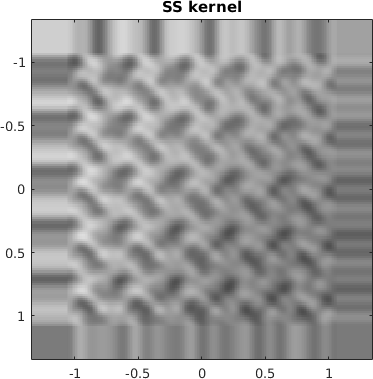}
    \caption{}
    \end{subfigure}
    \begin{subfigure}[b]{0.195\textwidth}
    \includegraphics[width=\textwidth]{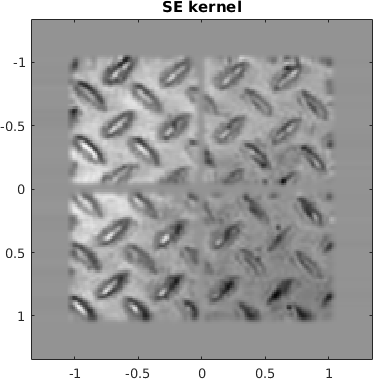}
    \caption{}
    \end{subfigure}
    
    \begin{subfigure}[b]{0.195\textwidth}
    \includegraphics[width=\textwidth]{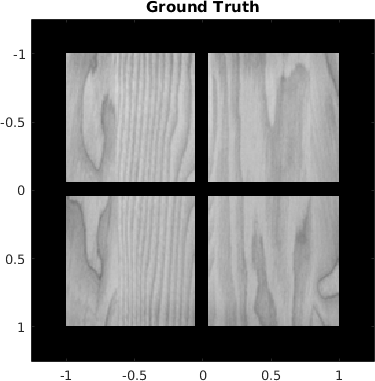}
    \caption{}
    \end{subfigure}
    \begin{subfigure}[b]{0.195\textwidth}
    \includegraphics[width=\textwidth]{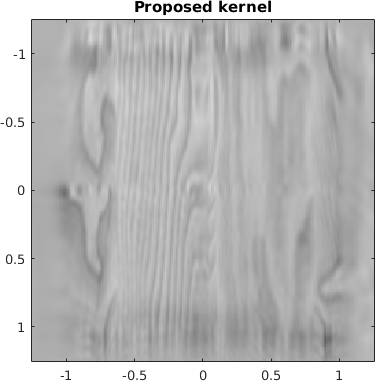}
    \caption{}
    \end{subfigure}
    \begin{subfigure}[b]{0.195\textwidth}
    \includegraphics[width=\textwidth]{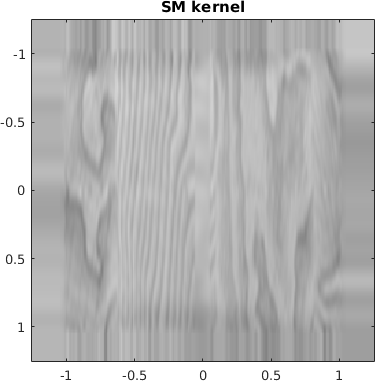}
    \caption{}
    \end{subfigure}
    \begin{subfigure}[b]{0.195\textwidth}
    \includegraphics[width=\textwidth]{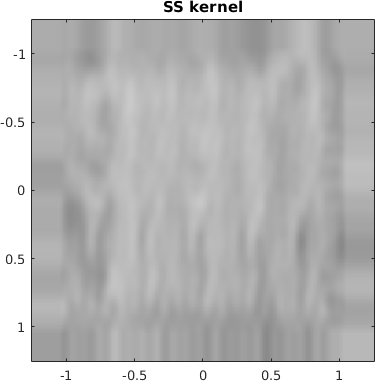}
    \caption{}
    \end{subfigure}
    \begin{subfigure}[b]{0.195\textwidth}
    \includegraphics[width=\textwidth]{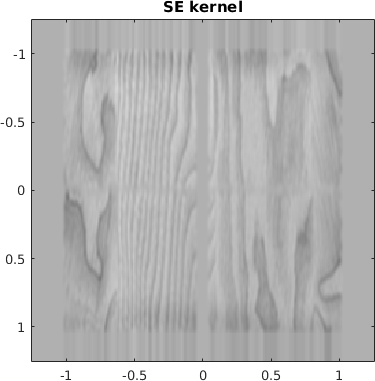}
    \caption{}
    \end{subfigure}
    \caption{A metal texture data with $Q=5$ components used for GSM, SM and SS kernels shown in \textbf{(a)}-\textbf{(e)} and a wood texture in \textbf{(f)}-\textbf{(j)} (with $Q=10$ components). The GSM kernel performs the best, making the most believable extrapolation outside image borders in \textbf{(b)} and \textbf{(g)}. The SM kernel fills in the missing cross pattern in \textbf{(c)} but does not extrapolate well. In \textbf{(h)} the SM kernel fills in the vertical middle block only with the mean value while GSM in \textbf{(g)} is able to fill in a wood-like pattern. SS is not able discover enough structure in either texture \textbf{(d)} or \textbf{(i)}, while the SE kernel overfits by using a too short length-scale in \textbf{(e)} and \textbf{(j)}.}
    \label{fig:metal}
\end{figure}

Figure \ref{fig:metal} shows the two texture images, and extrapolation predictions given by the proposed GSM kernel, with a comparison to the spectral mixture (SM), sparse spectrum (SS) and standard squared exponential (SE) kernels. For GSM, SM and SS we used $Q=5$ mixture components for the metal texture, and $Q=10$ components for the more complex wood texture. 

The GSM kernel gives the most pleasing result visually, and fills in both patterns well with consistent external extrapolation as well. The stationary SM kernel does capture the cross-section, but has trouble extrapolation outside the borders. The SS kernel fails to represent even the training data, it lacks any smoothness in the frequency space. The gaussian kernel extrapolates poorly.

\subsection{Spatio-Temporal Analysis of Land Surface Temperatures}

NASA\footnote{\url{https://neo.sci.gsfc.nasa.gov/view.php?datasetId=MOD11C1_M_LSTDA}} provides a land surface temperature dataset that we used to demonstrate our kernel in analysis of spatio-temporal data. Our primary objective is to demonstrate the capability of the kernel in inferring long-range, non-stationary spatial and temporal covariances.

We took a subset of four years (February 2000 to February 2004) of North American land temperatures for training data. In total we get 407,232 data points, constituting 48 monthly temperature measurements on a $84 \times 101$ map grid. The grid also contains water regions, which we imputed with the mean temperature of each month. We experimented with the data by learning a generalized spectral mixture kernel using $Q=5$ components. 

\begin{figure}
    \centering
    \begin{subfigure}[b]{0.49\textwidth}
    \begin{minipage}{\textwidth}
    \includegraphics[width=\textwidth]{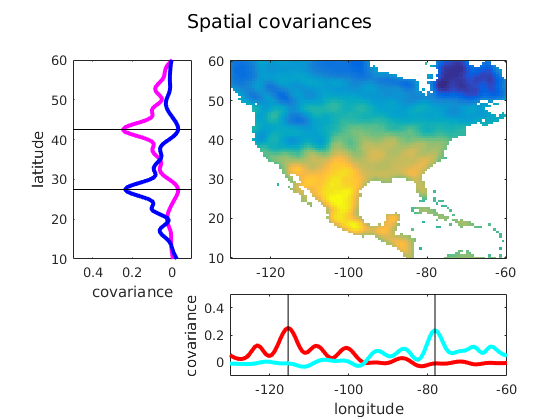}
    \end{minipage}
    \caption{}
    \end{subfigure}
    \begin{subfigure}[b]{0.49\textwidth}
    \begin{minipage}{\textwidth}
    \includegraphics[width=\textwidth]{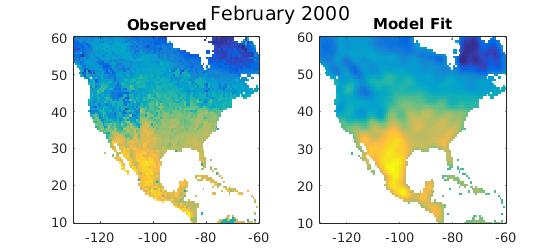}\\
    \includegraphics[width=\textwidth]{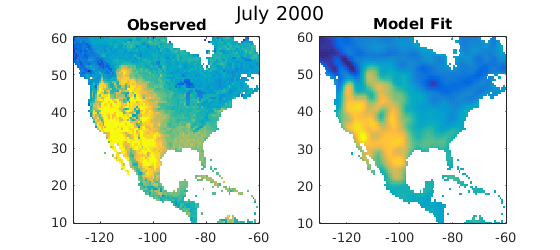}
    \end{minipage}
    \caption{}
    \end{subfigure}
    
    \begin{subfigure}[b]{\textwidth}
    \includegraphics[width=\textwidth]{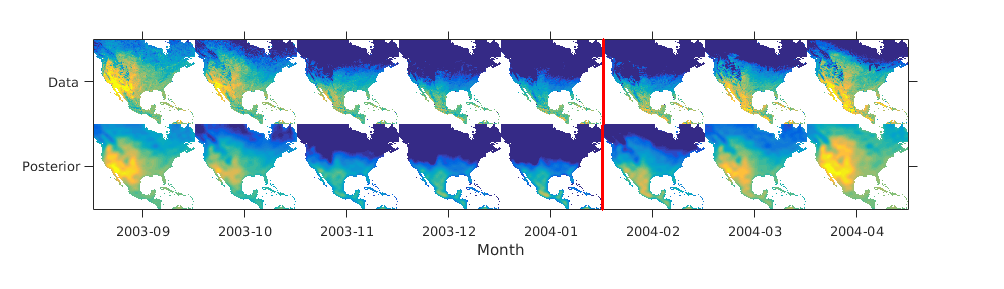}
    \caption{}
    \end{subfigure}
    
    \caption{\textbf{(a)} Demonstrates the non-stationary spatial covariances in the land surface data. The vertical black lines denote the point $x_0$ at which the kernel function $k(\cdot,x_0)$ is centered. \textbf{(b)} Sample reconstructions. In all plots, only the land area temperatures are shown. \textbf{(c)} Posterior for five last training months (until Jan 2004) and prediction for the three next months (February 2004 to April 2004), which the model is able to to construct reasonably accurately.}
    \label{fig:spatial}
\end{figure}

Figure~\ref{fig:spatial} presents our results. Figure~\ref{fig:spatial}b highlights the training data and model fits for a winter and summer month, respectively. Figure~\ref{fig:spatial}a shows the non-stationary kernel slices at two locations across both latitude and longitude, as well as indicating that the spatial covariances are remarkably non-symmetric. Figure~\ref{fig:spatial}c indicates five months of successive training data followed by three months of test data predictions. 

\section{Discussion}\label{sec:discussion}

In this paper we have introduced non-stationary spectral mixture kernels, with treatment based on the generalised Fourier transform of non-stationary functions. We first derived the bivariate spectral mixture (BSM) kernel as a mixture of non-stationary spectral components. However, we argue it has only limited practical use due to requiring an impractical amount of components to cover any sufficiently sized input space. The main contribution of the paper is the generalised spectral mixture (GSM) kernel with input-dependent Gaussian process frequency surfaces. The Gaussian process components can cover non-trivial input spaces with just a few interpretable components. The GSM kernel is a flexible, practical and efficient kernel that can learn both local and global correlations across the input domains in an input-dependent manner. We highlighted the capability of the kernel to find interesting patterns in the data by applying it on climate data where it is highly unrealistic to assume the same (stationary) covariance pattern for every spatial location irrespective of spatial structures.

Even though the proposed kernel is motivated by the generalised Fourier transform, the solution to its spectral surface 
\begin{align}
S_{\text{GSM}}(s,s') = \iint k_{\text{GSM}}(x,x') e^{-2 \pi i (xs - x's')} dx dx' \label{eq:gsmsurface}
\end{align}
remains unknown due to having multiple GP functions inside the integral. Figure~\ref{fig:surfaces}h highlights a numerical integration of the surface equation \eqref{eq:gsmsurface} on an example GP frequency surface. Furthermore, the theoretical work of Kom Samo and Roberts \cite{komsamo2015} on generalised spectral transforms suggests that the GSM kernel may also be dense in the family of non-stationary kernels, that is, to reproduce arbitrary non-stationary kernels.

\renewcommand\bibsection{\section*{\refname}}
\bibliographystyle{abbrv}
\bibliography{refs}

\appendix
\input{suppl}

\end{document}

%% file: math_macros.tex
\usepackage{amsmath,amssymb,amsthm}
\newcommand{\x}{\mathbf{x}}

\newcommand{\w}{\mathbf{w}}

\newcommand{\bl}{{\boldsymbol\ell}}

\newcommand{\y}{\mathbf{y}}

\newcommand{\f}{\mathbf{f}}

\newcommand{\K}{\mathbf{K}}
\renewcommand{\L}{\mathcal{L}}
\newcommand{\LL}{\mathbf{L}}
\newcommand{\E}{\mathbf{E}}
\newcommand{\bt}{{\boldsymbol{\theta}}}
\newcommand{\tbt}{{\boldsymbol{\tilde\theta}}}

\newcommand{\R}{\mathbb{R}}
\newcommand{\0}{\mathbf{0}}
\newcommand{\I}{\mathbf{I}}
\newcommand{\N}{\mathcal{N}}
\newcommand{\GP}{\mathcal{GP}}

\newcommand{\bmu}{\boldsymbol{\mu}}

\newcommand{\bb}{\mathbf{b}}

\newcommand{\half}{\frac{1}{2}}

\renewcommand{\vec}[1]{\boldsymbol{#1}}
\newcommand{\mat}[1]{\boldsymbol{#1}}

\DeclareMathOperator{\diag}{diag}

\DeclareMathOperator{\cov}{\textbf{cov}}

\DeclareMathOperator{\trace}{tr}
\DeclareMathOperator{\logit}{logit}

\newcommand{\pdd}[2]{\frac{\partial #1}{\partial #2}}

%% file: suppl.tex
\section{A tutorial on Gaussian processes}

We summarise here Gaussian process regression for completeness. For an interested reader, we refer to the excellent and comprehensive book by Rasmussen and Williams \cite{rasmussen2006}.

Gaussian processes (GP) are a Bayesian nonparameteric machine learning framework for regression, classification and unsupervised learning \cite{rasmussen2006}. A Gaussian process is a collection of random variables, any finite combination of which has a Multivariate normal distribution. A GP prior defines a distribution over functions, denoted as
\begin{align}
f(x) \sim \GP(m(x), k(x,x')),
\end{align}
where the mean function $m(x)$ and a positive semi-definite kernel function $K(x,x')$ for inputs $x \in \R$ determine the function expectation and covariance,
\begin{align}
\E [ f(x)] &= m(x) \\
\cov[f(x), f(x')] &= k(x,x').
\end{align}
Furthermore, the GP prior determines that for any finite collection of input points $x_1, \ldots, x_N$, the corresponding function values follow a Multivariate normal distribution
\begin{align}
p( f(x_1), \ldots, f(x_N) ) \sim \N( \mathbf{m}, K),
\end{align}
where $\mathbf{m} = (m(x_1), \ldots, m(x_N))^T \in \R^N$, and $K \in \R^{N \times N}$ with $K_{ij} = k(x_i, x_j)$. A Gaussian process models functions where for similar points $x,x'$ their corresponding function values $f(x),f(x')$ are also similar. A common kernel choice is the Gaussian kernel
\begin{align}
k(x,x') = \sigma_f^2 \exp\left(- \frac{1}{2} \frac{(x-x')^2}{\ell^2}\right),
\end{align}
which encodes monotonic neighborhood similarity. The kernel parameters are the signal variance $\sigma_f^2$ and the kernel lengthscale $\ell$.

Assume a dataset $\mathcal{D} = (x_i,y_i)_{i=1}^N$ and an additive Gaussian likelihood
\begin{align}
y = f(x) + \varepsilon(x), \qquad \varepsilon(x) \sim \N(0, \sigma_n^2)
\end{align}
with a data likelihood
\begin{align}
p(\y | \f) = \N(\y | \f, \sigma_n^2I),
\end{align}
where $\y = (y_1, \ldots, y_N)^T \in \R^N$ collects the observed outputs corresponding to inputs $(x_1, \ldots, x_N)$, and $\f = (f(x_1), \ldots, f(x_N))^T \in \R^N$ collects the function values, and $\sigma_n^2$ is the noise variance. The predictive distribution of $f(x_*) | \y$ for a new point $x_*$ conditioned on the data $\y$ at training inputs $X$ is again a Gaussian
\begin{align}
f(x_*) | \y &\sim \N(\mu_*, \sigma_*^2) \\
\mu_* &= K(x_*, X) (K + \sigma_n^2)^{-1} (\y - \mathbf{m}) + \mathbf{m} \\
\sigma_*^2 &= K(x_*,x_*) - K(x_*, X) ( K + \sigma_n^2)^{-1} K(X,x_*),
\end{align}
where $K(x_*,X) = K(X,x_*)^T$ is a row kernel.

Since the full predictive distribution is in closed form, the inference task is shifted to learning the hyperparameters $\theta = (\sigma_f, \ell, \sigma_n)$. The log marginalized likelihood 
\begin{align}
\log p(\y | \theta) &= \log \int p(\y | \f) p(\f | \theta) d\f \\
 &= \log \int \N(\y | \f, \sigma_n^2 I) \N(\f | \mathbf{m}, K) d\f \\
 &= \log \N(\y | \mathbf{m}, K + \sigma_n^2 I) \\
 &\propto - \frac{1}{2} (\y - \mathbf{m})^T (K + \sigma_n^2)^{-1} (\y - \mathbf{m}) - \frac{1}{2} \log | K + \sigma_n^2 |
\end{align}
has a closed form as well. The marginal log likelihood is related to the amount of functions compatible with the prior and matching the data. Hence, the marginal log likelihood automatically promotes priors that induce functions matching the data while penalising model complexity. The marginal log likelihood can be directly maximised using standard gradient ascent techniques to infer optimal hyperparameters $\theta$.

\section{Deriving the bivariate spectral mixture kernel}

A non-stationary kernel $k(x,x') \in \R$ for scalar inputs $x,x' \in \R$ can characterized by its spectral density $S(s,s')$ over frequencies $s,s' \in \R$, and the two are related via a generalised Fourier transform \cite{yaglom1987correlation,loeve1978probability}
\begin{align}
k(x,x') = \int_{\R} \int_{\R} e^{2\pi i (xs - x' s')} \mu_S(ds, ds') \label{eq:fourier}
\end{align}
where $\mu_S$ is a Lebesgue-Stieltjes measure associated to some positive semi-definite (PSD) spectral density function $S(s,s')$ with bounded variations, which we denote as the \emph{spectral surface} since it considers the amplitude of frequency pairs.

We define a spectral density $S(s,s')$ as a mixture of $Q$ bivariate Gaussian components
\begin{align}
  S_i(s,s') &= \sum_{ \bmu_i \in \pm \{ \mu_i, \mu_i'\}^2 } \N \left( \begin{pmatrix} s \\ s' \end{pmatrix} | \bmu_i, \Sigma_i\right) \label{eq:bsm} \\
  & \Sigma_i = \begin{bmatrix} \sigma_i^2 & \rho_i \sigma_i \sigma_i' \\ \rho_i \sigma_i \sigma_i' & {\sigma_i'}^2 \end{bmatrix} \notag
\end{align}
with parameterization using the correlation $\rho_i$, means $\mu_i,\mu_i'$ and variances $\sigma_i^2,{\sigma_i'}^2$. To ensure the PSD property of spectral density $S_i(s,s')$ it must hold that $S_i(s,s') = S_i(s',s)$ and sufficient diagonal components $S_i(s,s)$, $S_i(s',s')$ exist. In addition to retrieve a real-valued kernel we require symmetry with respect to the negative frequencies as well, i.e. $S_i(s,s') = S_i(-s,-s')$. The sum $\sum_{ \bmu_i \in \pm \{ \mu_i, \mu_i'\}^2 }$ satisfies all three requirements by iterating over four permutations of $\{\mu_i,\mu_i'\}^2$ and the opposite signs $(-\mu_i, -\mu_i')$, resulting in eight components \begin{align*}
\pm \{\mu,\mu'\}^2 = \{(\mu,\mu),(\mu,\mu'),(\mu',\mu),(\mu',\mu'),(-\mu,-\mu),(-\mu,-\mu'),(-\mu',-\mu),(-\mu',-\mu')\}.
\end{align*}
The full $Q$-component spectral density is
\begin{align}
  S(s,s') = \sum_{i=1}^Q \sum_{ \bmu_i \in \pm \{ \mu_i, \mu_i'\}^2 } \N \left( \begin{pmatrix} s \\ s' \end{pmatrix} | \bmu_i, \Sigma_i\right).
\end{align}

Next, we compute the generalised Fourier transform in closed form by exploiting Gaussian integral identities
\begin{align}
k(x,x') &= \int_\R \int_\R S(s,s') e^{2\pi i (xs - x' s')} ds ds' \\
 &= \int_{\R \times \R} \sum_{i=1}^Q \sum_{ \bmu_i \in \pm \{ \mu_i, \mu_i'\}^2 } \N \left( \begin{pmatrix} s \\ s' \end{pmatrix} | \bmu_i, \Sigma_i\right) e^{2 \pi i \tilde{\x}^T\mathbf{s}} d\mathbf{s} \\
 &= \sum_{i=1}^Q \sum_{ \bmu_i \in \pm \{ \mu_i, \mu_i'\}^2 } \int_{\R \times \R} \N ( \mathbf{s} | \bmu_i, \Sigma_i ) e^{2 \pi i \tilde{\x}^T\mathbf{s}} d\mathbf{s} \\
 &=  \sum_{i=1}^Q \sum_{ \bmu_i \in \pm \{ \mu_i, \mu_i'\}^2 } \frac{1}{(2\pi)^2 | \Sigma_i |} \int \exp\left( -\frac{1}{2} (\mathbf{s}- \bmu_i)^T \Sigma_i^{-1} (\mathbf{s}- \bmu_i)  + \mathbf{b}^T \mathbf{s}\right) d\mathbf{s} \\
 &= \sum_{i=1}^Q \sum_{ \bmu_i \in \pm \{ \mu_i, \mu_i'\}^2 } \frac{w^2}{(2\pi)^2 | \Sigma_i |} \int \exp\left( -\frac{1}{2} \mathbf{s}^T \Sigma_i^{-1} \mathbf{s}  + (\mathbf{b} +  \Sigma_i^{-1}\bmu_i )^T \mathbf{s} - \frac{1}{2} \bmu_i^T \Sigma_i^{-1}  \bmu_i \right) d\mathbf{s} \\
 &=  \sum_{i=1}^Q \sum_{ \bmu_i \in \pm \{ \mu_i, \mu_i'\}^2 } \exp\left( \frac{1}{2} (\mathbf{b} +  \Sigma_i^{-1} \bmu_i)^T \Sigma_i (\mathbf{b} +  \Sigma_i^{-1} \bmu_i ) \right) \exp \left(- \frac{1}{2} \bmu_i^T \Sigma_i^{-1}  \bmu_i  \right) \\
 &=  \sum_{i=1}^Q \sum_{ \bmu_i \in \pm \{ \mu_i, \mu_i'\}^2 } \exp\left( \frac{1}{2} \bb^T \Sigma_i \bb + \bmu_i^T \bb \right)
\end{align}
where we defined $\tilde{\x} = ( x, -x')^T$ and $\mathbf{s} = (s, s')^T$, and $\mathbf{b} = ( 2 \pi i x, -2 \pi i x')^T$. 

The $i$'th component of the kernel mixture is then
\begin{align}
    k_i(x,x') &= e^{-2\pi^2\tilde{\x}^T\Sigma \tilde{\x}} [
        & e^{2\pi i \mu x}e^{-2\pi i \mu' x'} + e^{2\pi i \mu' x}e^{-2\pi i \mu x'} + e^{2\pi i \mu x}e^{-2\pi i \mu x'} + e^{2\pi i \mu' x}e^{-2\pi i \mu' x'} \label{eq:sum8}\\
        &&+ e^{-2\pi i \mu x}e^{2\pi i \mu' x'} + e^{-2\pi i \mu' x}e^{2\pi i \mu x'} + e^{-2\pi i \mu x}e^{2\pi i \mu x'} + e^{-2\pi i \mu' x}e^{2\pi i \mu' x'} ] \nonumber
\end{align}
which can be simplified by noting that
\begin{align*}
    &e^{2\pi i \mu x}e^{-2\pi i \mu' x'} + e^{-2\pi i \mu x}e^{2\pi i \mu' x'} \\
    &= (\cos(2\pi \mu x) + i\sin(2\pi \mu x))(\cos(2\pi \mu' x') - i\sin(2\pi \mu' x')) \\
    &+ (\cos(2\pi \mu x) - i\sin(2\pi \mu x))(\cos(2\pi \mu' x') + i\sin(2\pi \mu' x')) \\
    &= 2\cos(2\pi \mu x) \cos(2\pi \mu' x') + 2\sin(2\pi \mu x)\sin(2\pi \mu' x')
\end{align*}
where the complex part cancels out. Now by defining a function
\begin{align}
    \Psi_{\mu,\mu'}(x) = \begin{pmatrix} \cos 2\pi\mu x + \cos 2\pi\mu' x \\ \sin 2\pi\mu x + \sin 2\pi\mu' x \end{pmatrix} \label{eq:psi}
\end{align}
we can express the sum of the 8 exponentials in \eqref{eq:sum8} as $\Psi_{\mu,\mu'}(x)^T \Psi_{\mu,\mu'}(x')$. The final kernel thus takes the form
\begin{align}
     k(x,x') = \sum_{i=1}^Q w_i^2 e^{-2\pi^2 \tilde{\x}^T\Sigma_i \tilde{\x}} \Psi_{\mu_i,\mu'_i}(x)^T\Psi_{\mu_i,\mu'_i}(x'), \label{eq:nonstat}
\end{align}
where we introduced mixture weights $w_i$ for each component. 

Now, we immediately notice that the kernel vanishes rapidly outside the origin $(x,x') = (0,0)$; we would require a huge number of components centered at different points $x_i$ to cover a reasonably-sized input space. One simple fix would be to change the exponential part to e.g. a Gaussian kernel $\exp(-\half \sigma^2||x-x'||^2)$ to prevent the component from vanishing but this still would not allow us to account for non-stationary frequencies, which is what we address next.

\section{Deriving the generalised spectral mixture (GSM) kernel}

The generalised spectral mixture kernel defines Gaussian process frequencies, lengthscales and mixture weights:
\begin{align}
    \log w_i(x) \sim \GP(0,k_w(x,x')), \\
    \log \ell_i(x) \sim \GP(0,k_\ell(x,x')), \\
    \logit \mu_i(x) \sim \GP(0,k_\mu(x,x')),
\end{align}
where we use the log transform to ensure weights $w(x)$ and lengthscales $\ell(x)$ are positive, and we use the logit transformed. The transform $\hat\mu$ and the inverse transform $\mu$ is given by 
\begin{align}
\logit(\mu) = \hat\mu &= \log\frac{\mu}{F_N-\mu} \\ 
\mu &= \frac{F_N}{1+\exp(-\hat\mu)}.
\end{align}
Frequency parameter $\logit \mu(x)$ to limit the learned frequencies between zero and the Nyquist frequency $F_N$, which can be defined as half of the sampling rate of the signal (or for non-equispaced signals as the inverse of the smallest time interval between the samples).

To accommodate lengthscale functions we replace the exponential part of the BSM kernel by the Gibbs kernel
\begin{align*}
    k_{gibbs,i}(x,x') = \sqrt{\frac{2\ell_i(x) \ell_i(x')}{\ell_i(x)^2+\ell_i(x')^2}}\exp\left(-\frac{(x-x')^2}{\ell_i(x)^2+\ell_i(x')^2}\right).
\end{align*}
The cosine part \eqref{eq:psi} is replaced by a function
\begin{align*}
\Psi_i(x) = \begin{pmatrix} \cos(2\pi\mu_i(x)x) \\ \sin(2\pi\mu_i(x)x) \end{pmatrix}.
\end{align*}
The non-stationary \emph{generalised spectral mixture} (GSM) kernel has a closed form
\begin{align}
k_{gsm}(x,x') &= \sum_{i=1}^Q w_i(x)w_i(x') k_{gibbs}(x,x') \Psi_i(x)^T \Psi_i(x') \\
 &= \sum_{i=1}^Q w_i(x) w_i(x') k_{gibbs,i}(x,x') \cos(2 \pi (\mu_i(x) x - \mu_i(x') x'))
\end{align}
due to identity $\cos \alpha \cos \beta + \sin \alpha \sin \beta = \cos(\alpha - \beta)$. The kernel is a product of three kernels, namely a linear kernel, a Gibbs kernel and a novel cosine kernel with a feature mapping $\Psi_i(x)$. The full kernel is PSD due to all of its product kernels being PSD. The cosine kernel is PSD due to a dot product.

\subsection{Relationship between Spectral Mixture kernel and the Generalised Spectral Mixture kernel}

We show that the proposed non-stationary GSM kernel reduces to the stationary SM kernel with appropriate parameterisation. We show this identity for univariate inputs for simplicity, with the same result being straightforward to derive for multivariate kernel variants as well.

The proposed generalised spectral mixture (GSM) kernel for univariate inputs is
\begin{align}
k_{\text{GSM}}(x,x') = \sum_{i=1}^Q w_i(x) w_i(x') \sqrt{ \frac{2 \ell_i(x) \ell_i(x') }{\ell_i(x)^2 + \ell_i(x')^2} } \exp\left(- \frac{(x-x')^2}{\ell_i(x)^2 + \ell_i(x')^2} \right) \cos\left(2 \pi (\mu_i(x) x - \mu_i(x') x')\right)
\end{align}
with Gaussian process functions $w_i(x), \mu_i(x), \ell_i(x)$. The Spectral Mixture (SM) kernel by Wilson et al~\cite{wilson2013} is
\begin{align}
k_{\text{SM}}(x,x') &= \sum_{i=1}^Q w_i^2  \exp( -2 \pi^2 (x-x')^2 \sigma_i^2) \cos( 2 \pi \mu_i (x-x')) \\
S_{\text{SM}}(s) &= \sum_{i=1}^Q w_i^2 \left[ \N( s | \mu_i, \sigma_i^2) + \N(s | -\mu_i, \sigma_i^2)\right],
\end{align}
where the parameters are the weights $w_i$, mean frequencies $\mu_i$ and variances $\sigma_i^2$. Now if we assign the following constant functions for the GSM kernel to match the parameters of the SM kernel on the right-hand side,
\begin{align}
w_i(x) &= w_i \\
\mu_i(x) &= \mu_i \\
\ell_i(x) &= \frac{1}{2 \pi \sigma_i},
\end{align}
we retrieve the SM kernel
\begin{align}
k_{\text{GSM}}(x,x') &= \sum_{i=1}^Q w_i(x) w_i(x') \sqrt{ \frac{2 \ell_i(x) \ell_i(x') }{\ell_i(x)^2 + \ell_i(x')^2} } \exp\left(- \frac{(x-x')^2}{\ell_i(x)^2 + \ell_i(x')^2} \right) \cos(2 \pi (\mu_i(x) x - \mu_i(x') x')) \\
 &= \sum_{i=1}^Q w_i^2 \exp\left(- \frac{(x-x')^2}{2 (1 / (2 \pi \sigma_i))^2} \right) \cos(2 \pi \mu (x - x')) \\
 &= \sum_{i=1}^Q w_i^2 \exp\left(- \frac{1}{2} (2 \pi \sigma_i)^2 (x-x')^2 \right) \cos(2 \pi \mu (x - x')) \\
 &= \sum_{i=1}^Q w_i^2 \exp\left(- 2 \pi^2 \sigma_i^2 (x-x')^2 \right) \cos(2 \pi \mu (x - x')) \\
 &= k_{\text{SM}}(x,x').
\end{align}
This indicates that the GSM kernel can reproduce any kernel that is reproducable by the SM kernel, which is known to be a highly flexible kernel \cite{wilson2013,wilson2015kissgp}. In practise we can simulate stationary kernels by setting the spectral function priors $k_w, k_\mu, k_\ell$ to enforce very smooth, or in practise constant, functions.

\section{Inference}

In many applications multivariate inputs have a grid-like structure, for instance in geostatistics, image analysis and temporal models. We exploit this assumption and propose a multivariate extension that assumes the inputs to decompose across input dimensions \cite{flaxman2015,wilson2013}:
\begin{align}
k_{\text{GSM}}(\x,\x' | \boldsymbol\theta) = \prod_{p=1}^P k_{\text{GSM}}(x_p, x_p' | \bt_p) \; .
\end{align}
Here $\x,\x' \in \R^P$, $\bt = (\bt_1, \ldots, \bt_P)$ collects the dimension-wise kernel parameters $\bt_p = (\w_{ip}, \bl_{ip}, \bmu_{ip})_{i=1}^Q$ of the $N$-dimensional realisations $\w_{ip},\bl_{ip},\bmu_{ip} \in \R^N$ per dimension $p$. Then, the kernel matrix can be expressed using Kronecker products as $\K_\bt = K_{\bt_1} \otimes \cdots \otimes K_{\bt_P}$, while missing values and data not on a regular grid can be handled with standard techniques \cite{flaxman2015,saatcci2011scalable,wilson2015kissgp}.

We use the Gaussian process regression framework and assume a Gaussian likelihood over $N^P$ data points $(\x_j,y_j)_{j=1}^{N^P}$ with all outputs collected into a vector $\y \in \R^{N^P}$,
\begin{align}
y_j &= f(\x_j) + \varepsilon_j, \qquad \varepsilon_j \sim \N(0, \sigma_n^2) \notag \\
 f(\x) &\sim \GP(0, k_{\text{GSM}}(\x,\x' | \bt)),
\end{align}
with a standard predictive GP posterior $f(\x_\star | \y)$ for a new input point $\x_\star$ \cite{rasmussen2006}. The posterior can be efficiently computed using Kronecker identities \cite{saatcci2011scalable}.

We aim to infer the noise variance $\sigma_n^2$ and the kernel parameters $\boldsymbol\theta = (\w_{ip}, \bl_{ip}, \bmu_{ip})_{i=1,p=1}^{Q,P}$ that reveal the input-dependent frequency-based correlation structures in the data, while regularising the learned kernel to penalise overfitting. We perform MAP inference over the log marginalized posterior $\log p(\bt | \y) \propto \log p(\y | \bt) p(\bt) = \mathcal{L}(\bt)$, where the functions $f(x)$ have been marginalised out,
\begin{align}
\mathcal{L}(\bt) &= \log \left( \N(\y | \0, \K_{\bt} + \sigma_n^2\I) \prod_{i,p=1}^{Q,P} \N(\w_{ip} | \0, K_{w_p}) \N(\bmu_{ip} | \0, K_{\mu_p}) \N(\bl_{ip} | \0, K_{\ell_p}) \right)  \label{eq:gpmll} \\ 
 &\propto - \y^T (\K_{\bt} + \sigma^2 I)^{-1} \y - \log | \K_{\bt} + \sigma_n^2 I | \notag \\
 &\quad- \sum_{p=1}^P \sum_{i=1}^Q \left( \w_{ip}^T K_{w_p}^{-1} \w_{ip} - \bl_{ip}^T K_{\ell_p}^{-1} \bl_{ip} - \bmu_{ip}^T K_{\mu_p}^{-1} \bmu_{ip} \right) - Q \sum_{p=1}^P \left(\log |K_{w_p}| - \log |K_{\ell_p}| -\log |K_{\mu_p} | \right) \notag
\end{align}
where $K_{w_p},K_{\mu_p},K_{\ell_p}$ are $N \times N$ prior matrices per dimensions $p$. The marginalized posterior automatically balances between parameters $\bt$ that fit the data and a model that is not overly complex \cite{rasmussen2006}. 
We can efficiently evaluate both the marginalized posterior and its gradients in $\mathcal{O}( P N^{\frac{P+1}{P}} )$ instead of the usual $\mathcal{O}( {N^P}^3 )$ complexity \cite{saatcci2011scalable} (See Supplements).

Gradient-based optimisation of \eqref{eq:gpmll} is likely to converge very slowly due to parameters $\w_{ip}, \bmu_{ip}, \bl_{ip}$ being highly self-correlated. We remove the correlations by whitening the variables as $\tilde\bt = \LL^{-1}\bt$ where $\LL$ is the Cholesky decomposition of the prior covariances. We maximize $\L(\bt)$ using gradient ascent with respect to the whitened variables $\tbt$ by evaluating $\L( \LL \tbt)$ and the gradient as \cite{kuss2005,heinonen2016}
\begin{align}
\pdd{\L(\bt)}{\tbt} = \pdd{\L(\bt)}{\bt}\pdd{\bt}{\tbt} = \LL^T \pdd{\L(\bt)}{\bt}. \end{align}

\subsection{Kronecker inference}

The marginal likelihood \eqref{eq:gpmll} can be evaluated using the eigen decomposition $\K = \mat{Q}\mat{V}\mat{Q}^T$. Using known results for Kronecker products we can compute the eigen decomposition as $\mat{Q} = \bigotimes_p {Q}_p$, $\mat{V} = \bigotimes_p {V}_p$ and $\mat{Q}^T = \bigotimes_p {Q}_p^T$ using the decompositions of the smaller kernels $K_p = {Q}_p {V}_p {Q}_p^T$. Thus we can decompose the computation of the first term in \eqref{eq:gpmll} as
\begin{align}
    (\K+\sigma_n^2\I)^{-1}\y = \mat{Q}(\mat{V} + \sigma_n^2\I)^{-1}\mat{Q}^T\y = \left(\bigotimes_p {Q}_p\right) \left( (\mat{V} + \sigma_n^2\I)^{-1} \left( \left(\bigotimes_p {Q}_p^T\right) \y \right) \right), \label{eq:alpha}
\end{align}
where the inversion is taken only of the diagonal matrix of eigenvalues and matrix-vector products with a Kronecker matrix can be computed efficiently. The second term of \eqref{eq:gpmll} can be computed using the eigenvalues $\vec\lambda = \diag(\mat{V}) = \bigotimes_p \diag({V}_p)$ as $\log |\K + \sigma_n^2\I| = \sum_i \log(\lambda_i + \sigma_n^2)$.

The gradient of the marginal likelihood is given by
\begin{align}
    \pdd{\L}{\theta_p} = \half \left( \vec\alpha^T \pdd{\K}{\theta_p} \vec\alpha - \trace\left( (\K+\sigma_n^2\I)^{-1}\pdd{\K}{\theta_p} \right) \right), \label{eq:grad}
\end{align}
where $\vec\alpha = (\K+\sigma_n^2\I)^{-1}\y$ is computed as in \eqref{eq:alpha}. The gradient of the Kronecker product kernel can be computed as
\begin{align}
    \pdd{\K}{\theta_p} = K_1 \otimes \ldots \otimes \pdd{K_p}{\theta_p} \otimes \ldots \otimes K_P
\end{align}
assuming that $\pdd{\K_p}{\theta_i} = \vec0$ for $i \neq p$. As this is a Kronecker product, the first term in \eqref{eq:grad} can be computed efficiently. The trace term in \eqref{eq:grad} can be computed by exploiting the cyclic property and the eigen decomposition as
\begin{align}
    \trace\left( (\K+\sigma_n^2\I)^{-1}\pdd{\K}{\theta_p} \right) %&=  \trace\left( (\mat{V}+\sigma_n^2\I)^{-1} \mat{Q}^T \pdd{\K}{\theta_p}\mat{Q} \right) \\
    &= \diag\left( (\mat{V}+\sigma_n^2\I)^{-1} \right)^T \diag\left( \mat{Q}^T \pdd{\K}{\theta_p}\mat{Q} \right),
\end{align}
where the latter term can be computed efficiently as
\begin{align}
    \mat{Q}^T \pdd{\K}{\theta_p}\mat{Q} = {Q}_1^T K_1 {Q}_1 \otimes \ldots \otimes {Q}_p^T \pdd{K_p}{\theta_p} {Q}_p \otimes \ldots \otimes {Q}_P^T K_P {Q}_P
\end{align}
and its diagonal as a Kronecker product of the diagonals of each factor in the product. For the noise parameter $\sigma_n$ we get $\pdd{(\K+\sigma_n^2\I)}{\log\sigma_n} = 2\sigma_n^2\I$ which makes both terms in \eqref{eq:grad} easy to compute.

Kronecker methods are also easily extensible for non-complete grids \cite{wilson2015kissgp} and non-Gaussian likelihoods \cite{flaxman2015}.